\documentclass{article}

\usepackage{iclr2017_conference,times}
\usepackage{hyperref}
\usepackage{url}

\usepackage[backend=biber]{biblatex}
\usepackage{graphicx}
\graphicspath{{./images/}}

\title{Gear Training: A new way to implement high-performance model-parallel training}
\author{Hao Dong, Shuai Li, Dongchang Xu, Yi Ren, Di Zhang \\
\texttt{\{haodong,voolc.li,dongchang.xu,hengrui.ry,di.zhang\}@alibaba-inc.com} \\
}

\begin{document}

\maketitle

\begin{abstract}
	The training of Deep Neural Networks usually needs tremendous computing resources. Therefore many deep models are trained in large cluster instead of single machine or GPU. Though major researchs at present try to run whole model on all machines by using asynchronous asynchronous stochastic gradient descent (ASGD)\cite{Revisit_SGD}, we present a new approach to train deep model parallely -- split the model and then seperately train different parts of it in different speed.
\end{abstract}

\section{Introduction}
Many state-of-the-art deep learning models are dense models, which means they have large mounts of computing-expensive operations, such as ResNet\cite{RESNET}, Inception-v4\cite{INCEPTION} for image processing and LSTM\cite{LSTM}, GRU\cite{GRU} for audio applications. But in mnay Internet Search and Advertisement senarios, the models are much more complicated, since it would be compositions of sparse sub-model and dense sub-model\cite{Heng-Tze}.

In the scenario of Internet Advertisement we use series of user's page views as training data to predict the most proper advertises to display for different users. A page view mainly consists of \textit{User ID}, \textit{URL} and \textit{Image ID}. There would be multiple \textit{Image ID} for one page view since a page may contains many images. The structure of page views are shown in Table~\ref{page-view-table}

\begingroup
\begin{table}[ht]
\renewcommand{\arraystretch}{1.5}
\label{page-view-table}
\begin{center}
\begin{tabular}{c|c|c}
	\textit{User ID} & \textit{URL} & \textit{Image ID} \\
	\hline
	101 & https://item.taobao.com/item.htm?\&id=565598911558 & 85 \\
	352 & https://item.taobao.com/item.htm?\&id=564719868057 & 73 \\
	521 & https://item.taobao.com/item.htm?\&id=195911142551 & 92 \\
	... & ... & ... \\
\end{tabular}
\caption{Sample of page Views}
\end{center}
\end{table}
\endgroup

There would be tremendous page views for just one day. The state-of-the-art CNN models for image processing are very computing-hungry, so it will cost a few days to train all these page views if we directly assemble raw images into every page view record.

\begin{figure}[ht]
\begin{center}
\includegraphics[scale=0.5]{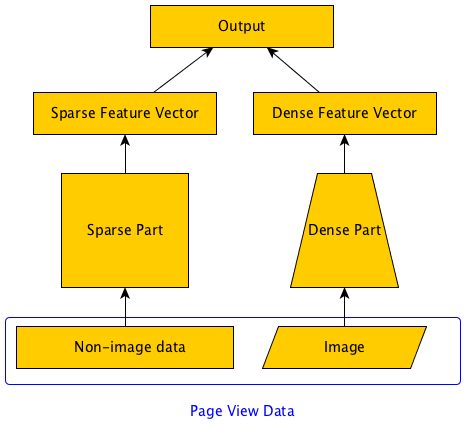}
\caption{Gear Training Model}
\label{fig:gear_training_model}
\end{center}
\end{figure}

To solve this problem, we present a new approach to train Spare-Dense-Combined Model: \textit{Gear Training}.
First, we split the model to two parts: \textit{Sparse Part} (for non-image data such as \textit{User ID}) and \textit{Dense Part} (for image, audio, video etc.). As shown in Figure \ref{fig:gear_training_model}. Then we can train different part on different GPU or different machines.

Second, it dosen't need to do inference work on \textit{Dense Part} of the model in every step of training. A training step consists of three phase: do inference through the model to get \textit{loss}; do back-propagation by using \textit{loss} to get \textit{gradients}; apply the \textit{gradients} on model. In inference phase, we store the inference results of \textit{Dense Part} as \textit{Dense Feature Vector} (in Figure \ref{fig:gear_training_model}) into KV store (HBase, in our production environment) by using \textit{Image ID} as key. Therefore, it only need to fetch inference results from KV store instead of computing inference of \textit{Dense Part} every time it gets a \textit{Image ID}. The inference results that stored in cache will be updated after a fixed period of time, which we call it \textit{inference expired time}.

Third, it don't need to do back-propagation work on \textit{Dense Part} in every step of training neither. The back-propagation results are also put into KV store by using \textit{Image ID} as key in another table. Different from inference process, which only need to store one inference result for a image, we store all gradients for one image in KV store. Only when a image's number of corresponding gradients has been accumulated to \textit{M}, it will average these gradients, and use the average consequence to do one step of back-progagation on \textit{Dense Part}.

To do back-propagation for \textit{Dense Part}, we need to get gradients of DFV (abbreviation for \textit{Dense Feature Vector}) in \ref{fig:gear_training_model} first:

\begin{equation}
\frac{\partial Loss} {\partial W_{Dense-Part}} = \frac{\partial Loss} {\partial DFV} \cdot \frac{\partial DFV} {\partial W_{Dense-part}}
\end{equation}

By doing inference every period of \textit{inference expired time} and back-propagation every \textit{M} steps of training on \textit{Dense Part}, it could save a lot of time or computing resources for training. The different training speeds on \textit{Dense Part} and \textit{Sparse Part} of model make them looks like different size of gears in a differential gear box. That is why we call our method 'Gear Training'.

\section{Related Works}
Distributed training of deep neural network has been widely used in industry. Many mainstream machine learning frameworks such as Tensorflow\cite{Tensorflow_MPI}, MXNET\cite{Mxnet_Dist}, BigDL\cite{BigDL} could support training models in clusters. Though training whole model with asynchronous stochastic gradient descent (ASGD)\cite{Revisit_SGD} is the most commonly used method for gradients update, there are already many different ways to split the model for distributing environment \cite{AMPNet} \cite{Data_Part}. And a bunch of relevant analysis has been published: \cite{Scala_Limits} \cite{SGD_paral}.

In paper of Daniel\cite{Daniel}, it average parameters of model to accelerate the speed of training in cluster. But there isn't any researchs to use averaged-gradients for back-propagation in DNN model yet.

\section{Method}
\begin{figure}[ht]
\begin{center}
\includegraphics[scale=0.3]{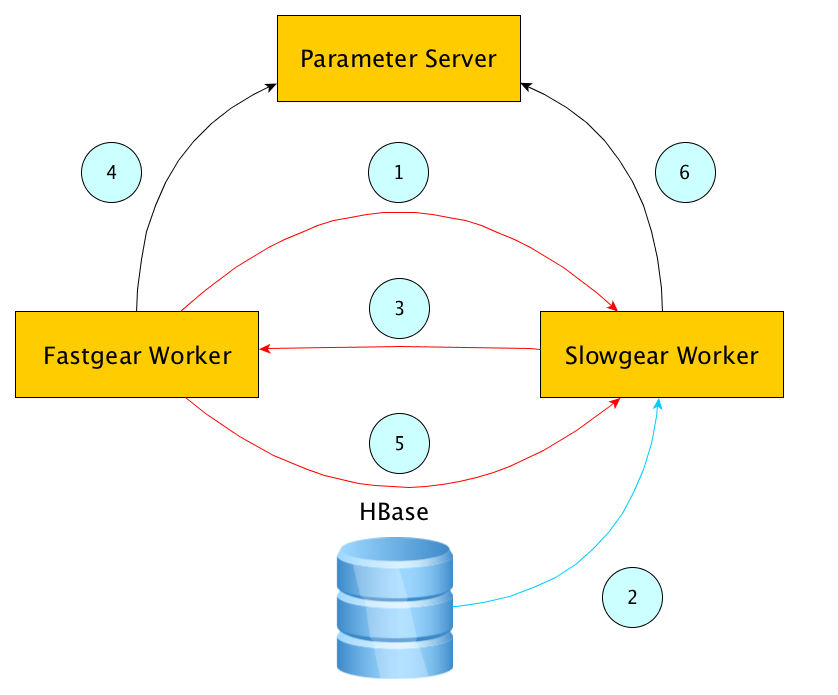}
\caption{Gear Training Sequence Diagram}
\label{fig:gear_diagram}
\end{center}
\end{figure}

We build the 'Gear Training' solution on Tensorflow Clusters by using our own Parameter Servers. There are two types of worker roles in cluster: \textit{Fastgear Worker} and \textit{Slowgear Worker}.
\textit{Fastgear Worker} is responsible for training \textit{Sparse Part}. Since computing of \textit{Sparse Part} is faster than \textit{Dense Part}, we call it 'Fastgear'. The basic working progress is shown in Figure \ref{fig:gear_diagram}

The detailed sequence events are explained below:

\begin{enumerate}
	\item \textit{Fastgear Worker} sends \textit{Image ID} to \textit{Slowgear Worker} to get inference result of \textit{Dense Part}.
	\item \textit{Slowgear Worker} gets raw image data from KV store by \textit{Image ID}.
	\item After doing inference on image, \textit{Slowgear Worker} sends inference result of this image back to \textit{Fastgear Worker}.
	\item Now with inference result of \textit{Dense Part}, \textit{Fastgear Worker} could do inference on \textit{Sparse Part} to get \textit{loss} and then do Back-propagation on \textit{Sparse Part}. After this, \textit{Fastgear Worker} will push gradients of \textit{Sparse Part} back to Parameter Servers. Noticed that Fastgear Worker would not do Back-propagation on \textit{Dense Part}. It only compute out the gradients to \textit{Dense Feature Vector}.
	\item \textit{Fastgear Worker} sends \textit{Image ID} and its corresponding gradients in \textit{Dense Feature Vector} to \textit{Slowgear Worker}.
	\item After getting \textit{M} gradients of \textit{Dense Feature Vector} for a image, \textit{Slowgear Worker} would average these gradients and using the consequence to do Back-propagation on \textit{Dense Part}. Finally, \textit{Slowgear Worker} push gradients of \textit{Dense Part} back to Parameter Servers.
\end{enumerate}

All raw images are stored in KV Store by using \textit{Image ID} as key. When training begin, it will launch a bunch of Fastgear Workers and Slowgear Workers. The Slowgear Workers will process requests from Fastgear Workers by partition of\textit{Image ID}, so we also cache raw images and its inference results in Slowgear Workers to accelerating computing. Unlike raw image which will not be changed at entire training stage, inference results are updated constantly according to the model. Therefore Slowgear Workers will drop expired inference results in cache continuously too.

\begin{figure}[h]
\begin{center}
\includegraphics[scale=0.3]{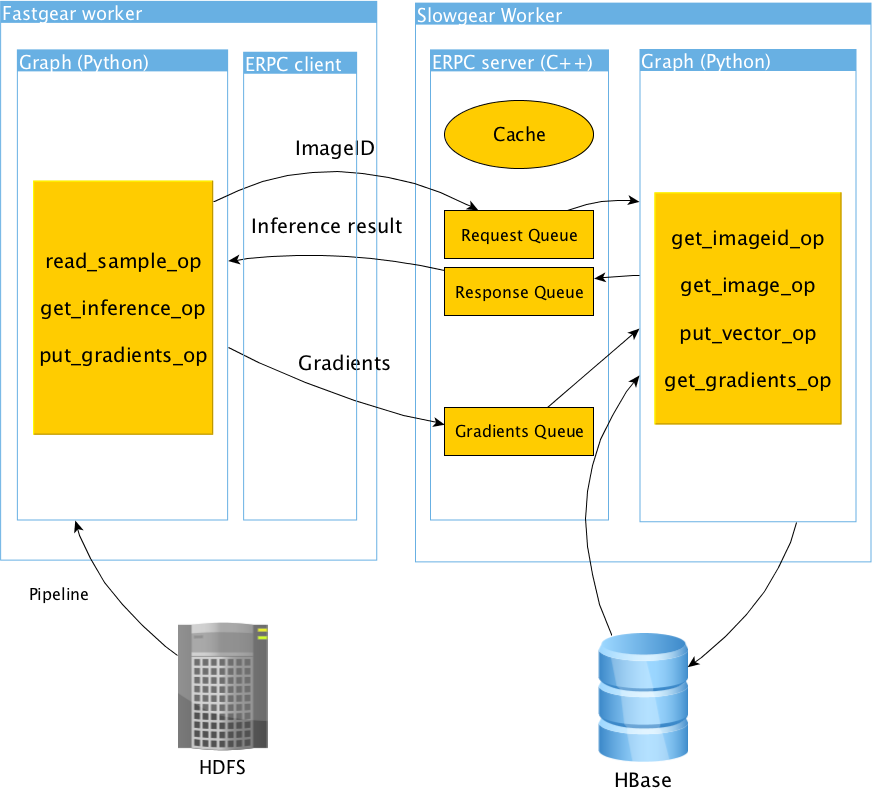}
\caption{Gear Trainig Architecture}
\label{fig:gear_arch}
\end{center}
\end{figure}

The software architecture of 'Gear Training' is shown in Figure \ref{fig:gear_arch}. The communications between Fastgear Workers and Slowgear Workers are implements by a sophisticated RPC framework in Alibaba which named 'ERPC'. All the \textit{send/recv/fetch-raw-image} primitives are implement as 'Operations' in Tensorflow, so customers could effortlessly write deep learning graph by using Tensorflow and add these 'Operations' in it to let model running as 'Gear Training' mode.

\section{Experiments}

\subsection{Experiments on CIFAR100}

\begin{figure}[h]
\begin{center}
\includegraphics[scale=0.4]{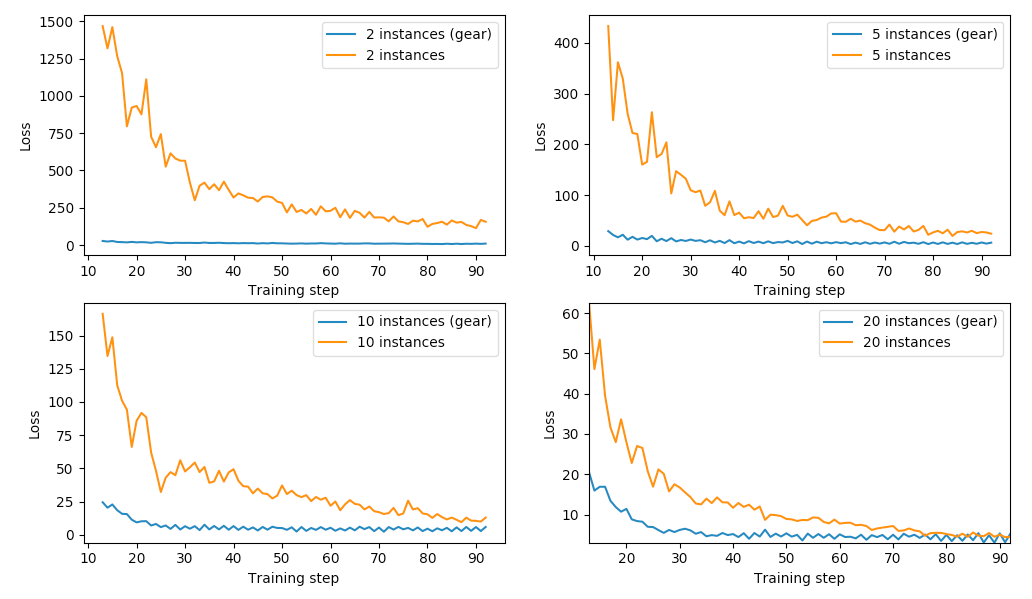}
\caption{Comparison between Gear Training and Traditional Training(nogear)}
\label{fig:gear_nogear_compare}
\end{center}
\end{figure}

\begin{figure}[h]
\begin{center}
\includegraphics[scale=0.4]{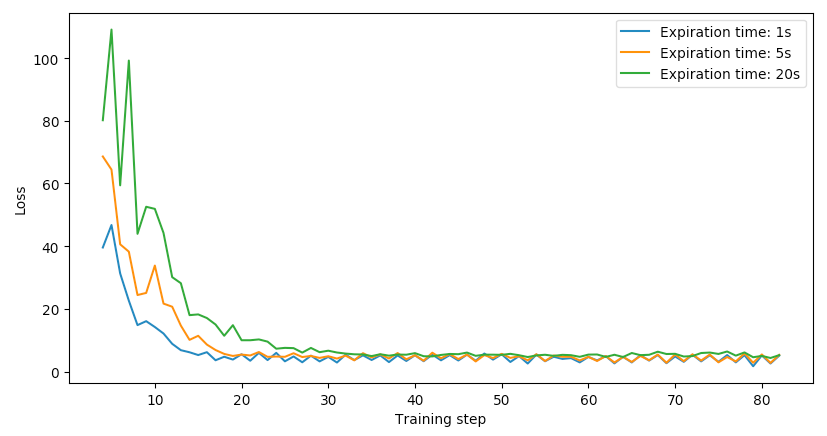}
\caption{Gear Training with different expire time for inference cache}
\label{fig:different_expire_inf_time}
\end{center}
\end{figure}

To verify that 'Gear Training' do improve the performance of training, we use it to train standard open dataset - CIFAR100\cite{CIFAR100}. We deliberately seperate every 32x32 image in CIFAR100 to two sub-images: top-half sub-image, and bottom-half sub-image. Then put top-half sub-image into \textit{Fastgear Worker}, and bottom-half sub-image into \textit{Slowgear Worker}, respectively.

We run different number of instances in our training cluster. In conventional training mode, we just use AdamOptimizer by setting learing rate to 1e-4. The initialization value's standard deviation of all variables are set to 0.1 since that too big standard deviation will cause training accuracy unable to converge. We run gear training mode with the same optimizer and initial configurations. In gear training mode we already have split the raw image, therefore the number of Slowgear Workers (and Fastgear Workers) should be equal to number of conventional training workers. For example, if we run 20 workers in conventional mode, it needs to launch 20 Slowgear Workers and 20 Fastgear Workers as well to complete the comparison test.
The testing consequences are shown in Figure \ref{fig:gear_nogear_compare}. The Gear Training method converged faster than conventional training method.

We also run 20 instances of Slowgear Workers and Fastgear Workers to see the different consequences by using different expired-time of inference cache, which is implemented in Slowgear Worker. As we can see in Figure \ref{fig:different_expire_inf_time}, longer expired-time will cause model to spend longer time for converging. But by looking at the Table~\ref{expire-performance-table}, we can see that longer expired-time also boosts the computing performance of training. Therefore in production environment, it needs to find the best expire time for inference cache on different models.

\begin{table}[h]
\renewcommand{\arraystretch}{1.5}
\label{expire-performance-table}
\begin{center}
\begin{tabular}{c|c}
	Expire time of inference cache (seconds) &  Times by training 500 steps (seconds) \\
	\hline
	1 & 282 \\
	5 & 276 \\
	20 & 233 \\
\end{tabular}
\caption{Times spend on training}
\end{center}
\end{table}

\subsection{Experiments on real application}

\begin{figure}[h]
\begin{center}
\includegraphics[scale=0.6]{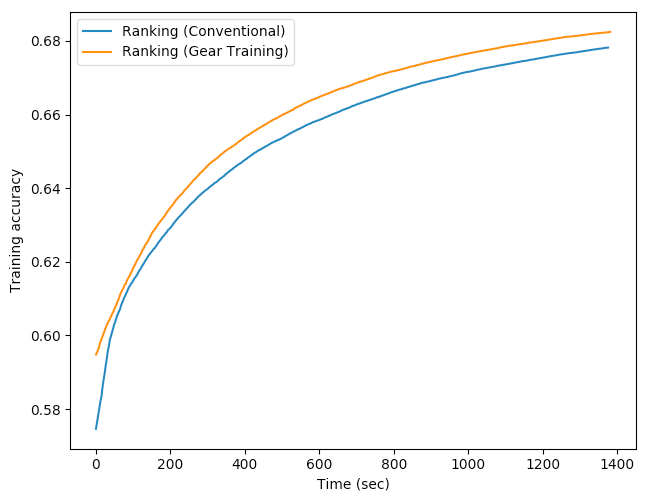}
\caption{Comparison between Gear Training and Traditional Training on 'Ranking'}
\label{fig:ranking}
\end{center}
\end{figure}

In advertisement department, we use a greate deal of page-views to predict the custom behaviors. We choose a real application called 'Ranking' to test the new Gear Training mode. By running 20 instances of workers with 60 seconds as expired-time for inference cache, we get the Figure \ref{fig:ranking}.

At the beginning stage, the training accuracy of gear training is near to the conventional method because the cache of inference. But after that, the change of parameters become more and more trival, hence the gear training's accuracy start to go beyond conventional method.

\section{Conclusion}
By splitting the deep neural network model and training it on different devices with different speeds, we can accelerate the speed of whole training process. The new method 'Gear Training' could boost the performance much higher in scenario which the number of images is much smaller than the number of total training samples.

\printbibliography

\end{document}